\documentclass[journal]{IEEEtran}
\ifCLASSINFOpdf
   \usepackage[pdftex]{graphicx}
   \graphicspath{{../pdf/}{../jpeg/}}
   \DeclareGraphicsExtensions{.pdf,.jpeg,.png}
\else
   \usepackage[dvips]{graphicx}
   \graphicspath{{../eps/}}
   \DeclareGraphicsExtensions{.eps}
\fi
\usepackage[tight,footnotesize]{subfigure}
\begin{document}
%
\title{Multiview Hessian Regularization for Image Annotation}
%
%
%

\author{Weifeng~Liu,~\IEEEmembership{Member,~IEEE}
        and~Dacheng~Tao,~\IEEEmembership{Senior Member,~IEEE}
\thanks{This work was supported in part by the Australian Research Council Discovery Project under Grant ARC DP-120103730, the National Natural Science Foundation of China under Grant 61271407, Shandong Provincial Natural Science Foundation, China under Grant ZR2011FQ016, the Fundamental Research Funds for the Central Universities, China University of Petroleum (East China) under Grant 13CX02096A. 

W. Liu is with the College of Information and Control Engineering, China University of Petroleum (East China), Qingdao 266580, China (e-mail: liuwf@upc.edu.cn).

D. Tao is with the Centre for Quantum Computation \& Intelligent Systems and the Faculty of Engineering and Information Technology, University of Technology, Sydney, Ultimo, NSW 2007, Australia (e-mail: dacheng.tao@uts.edu.au).

\copyright 20XX IEEE. Personal use of this material is permitted. Permission from IEEE must be obtained for all other uses, in any current or future media, including reprinting/republishing this material for advertising or promotional purposes, creating new collective works, for resale or redistribution to servers or lists, or reuse of any copyrighted component of this work in other works.
}
}

\maketitle

\begin{abstract}
The rapid development of computer hardware and Internet technology makes large scale data dependent models computationally tractable, and opens a bright avenue for annotating images through innovative machine learning algorithms. Semi-supervised learning (SSL) has consequently received intensive attention in recent years and has been successfully deployed in image annotation. One representative work in SSL is Laplacian regularization (LR), which smoothes the conditional distribution for classification along the manifold encoded in the graph Laplacian, however, it has been observed that LR biases the classification function towards a constant function which possibly results in poor generalization. In addition, LR is developed to handle uniformly distributed data (or single view data), although instances or objects, such as images and videos, are usually represented by multiview features, such as color, shape and texture. In this paper, we present multiview Hessian regularization (mHR) to address the above two problems in LR-based image annotation. In particular, mHR optimally combines multiple Hessian regularizations, each of which is obtained from a particular view of instances, and steers the classification function which varies linearly along the data manifold. We apply mHR to kernel least squares and support vector machines as two examples for image annotation. Extensive experiments on the PASCAL VOC'07 dataset validate the effectiveness of mHR by comparing it with baseline algorithms, including LR and HR.
\end{abstract}


%
\IEEEpeerreviewmaketitle

\section{Introduction}
%
%
%
%
\IEEEPARstart{T}{he} prodigious development of the digital camera, computer hardware, Internet technology, and machine learning technologies makes millions or even billions of images accessible to people. Large scale image annotation has therefore become essential for many practical applications in image processing, computer vision and multimedia. Since it is expensive to label a large number of images to train a robust learning model for image annotation, semi-supervised learning (SSL) has been introduced to semi-automatic image annotation and exploits both a small number of labeled images and a large number of unlabeled images to improve the generalization ability of a learning model.

The most common class of methods for SSL is based on the manifold assumption \cite{Belkin2006}, that is, two examples with similar features tend to share the same class label. Manifold regularization tries to explore the geometry of the intrinsic data probability distribution by penalizing the regression function along the potential manifold. Laplacian regularization (LR) \cite{Belkin2001,Belkin2006} is one of the representative works in which the geometry of the underlying manifold is determined by the graph Laplacian. LR-based SSL has received intensive attention and many algorithms have been developed, such as Laplacian regularized least squares (LapLS) and Laplacian support vector machines (LapSVM) \cite{Belkin2006}.

Although LR has shown its effectiveness in SSL, it is short of extrapolating power. The null space of the graph Laplacian is a constant function along the compact support of the marginal distribution and thus the solution of the LR is biased towards a constant function. This means that the function whose values are beyond the range of the training outputs is always a constant function, which causes LR-based SSL particularly to suffer when there are only few labeled examples.

In contrast to Laplacian, Hessian can properly exploit the intrinsic local geometry of the data manifold. Hessian has a richer nullspace and drives the learned function which varies linearly along the underlying manifold \cite{Eells1983,Donoho2003,Steinke2008,Kim}. Thus Hessian regularization (HR) cannot only properly fit the data within the domain defined by training samples, but it can also nicely predict the data points beyond the boundary of the domain. Compared to LR, HR is preferable for SSL for encoding the local geometry of the data distribution and thus can boost the SSL performance. In \cite{Kim}, the effectiveness of HR has been demonstrated for kernel regression.

The aforementioned learning methods, however, are only applicable to data represented by single view features, whereas in image annotation, images are naturally represented by multiview features, such as color, shape and texture. Each view of a feature summarizes a specific characteristic of the image, and features for different views are complementary to one another. Although we can concatenate different features into a long vector, this concatenation strategy (1) improperly treats different features carrying different physical characteristics, and (2) results in an over-fitting problem when the size of the training set is small. Therefore compared to single view learning, multiview learning can significantly improve performance especially when the weaknesses of one view can be reduced by the strengths of others.

In this paper, we present multiview Hessian regularization (mHR) for image annotation. Significantly, mHR optimally combines multiview features and Hessian regularizations obtained from different views. The advantages of mHR lie in the fact that: (1) mHR can steer the learned function which varies linearly along the underlying manifold and then extrapolate unseen data well; and (2) mHR can effectively explore the complementary properties of different features from different views and thus boost the image annotation performance significantly. We introduce mHR to kernel least squares and support vector machines for image annotation and conduct experiments on the PASCAL VOC'07 dataset \cite{EveringhamL2007}. To evaluate the performance of mHR, we also compare mHR with a number of baseline algorithms including Hessian SVM, Laplacian SVM, Hessian least squares, Laplacian least squares. The experimental results demonstrate the effectiveness of mHR by comparison with the baseline algorithms.

The rest of the paper is organized as follows. In section 2, we survey related work on SSL and multiview learning. Section 3 presents the proposed mHR framework and Section 4 details the implementation of mHR for least squares and support vector machines. Experimental results are detailed in Section 5, followed by the conclusion in Section 6.

\section{Related Work}
The proposed mHR framework is motivated by manifold regularization-based SSL \cite{Donoho2003,Belkin2006,Kim} and graph ensemble-based multiview learning \cite{Xia2010a,Xie2011}. This section briefly reviews the related works for better understanding mHR.

\subsection{SSL: Semi-supervised Learning}
In recent years, many SSL algorithms have been developed, and these algorithms can be grouped into the following three categories: generative models \cite{Miller1996,KamalNi,Fujino2005,caoliangliang2010a}, transductive support vector machines (TSVM) \cite{Vapnic1998} and graph-based methods \cite{Belkin2006}.

Generative models assume that examples are clustered and generated by the same parametric model. The purpose of these models is to learn the missing model parameters by employing the expectation-maximization algorithm \cite{Dempster1977}. Some typical models are mixture of Gaussian \cite{caoliangliang2010a}, mixture of experts \cite{Miller1996} and Na\"{\i}ve Bayes \cite{KamalNi,Tian2008}, \emph{etc.}

TSVM \cite{Vapnic1998} assigns potential labels to unlabeled data by initiating SVM using labeled examples; hence, in TSVM, a linear boundary has the maximum margin on both the labeled data and the unlabeled data. Unlabeled data guide the linear boundary traverse through the low density regions. To tackle the non-convexity of the loss function and computation complexity of TSVM, various methods have been proposed \cite{Chapelle2005,Collobert2006}.

Graph-based models assume that labels smooth over the graph and these models define a graph over training examples (labeled and unlabeled) to encode the similarity between examples. According to the regularization framework \cite{Belkin2006}, there are two terms in these methods: a loss function and a regularizer. By defining different loss functions and regularizers, a dozen methods can be obtained \cite{Elmoataz2008,Wangmeng2007}. Essentially, the graph regularizer plays a critical role and affects classification performance \cite{Zhu2007}.

\subsection{Multiview Learning}
Popular multiview learning algorithms can be grouped into the following three categories: co-training, multiple kernel learning and graph ensemble learning.

Co-training based algorithms learn from two different views \cite{Blum1998} which assumes that the features obtained from the two different views are sufficient to train a good classifier. Co-training methods effectively exploit unlabeled data to improve the classification performance, especially when the two views are conditionally independent of one another \cite{Nigam2000}. Several extensions of co-training have been proposed in recent years, such as co-EM \cite{Jones2005}, SVM-2K \cite{Farquhar} and co-SVM \cite{Brefeld2006}. Theoretical justifications can be found in \cite{Dasgupta2001}.

Multiple kernel learning (MKL) algorithms learn a kernel machine from multiple Gram kernel matrices \cite{Lanckriet2004,Bach2004,Sonnenburg2006,McFee2011,Meh2011}, each of which is built from a particular view of a feature. Lanckriet \emph{et al.} applied semidefinite programming (SDP) to MKL \cite{Lanckriet2004}. Bach \emph{et al.} \cite{Bach2004} applied sequential minimal optimization (SMO) to MKL and made MKL applicable to large scale data analytics problems. Sonnenburg \emph{et al.} \cite{Sonnenburg2006} reformulated the binary classification MKL problem as a semi-infinite linear programming problem. Gonen \emph{et al.} \cite{Meh2011} compared different MKL algorithms by performing experiments on real-world datasets.

Graph ensemble-based multiview learning algorithms integrate multiple graphs to explore the complementary properties of different views, each of which encodes the local geometry of a particular view. For example, Xia \emph{et al.} \cite{Xia2010a} developed multiview spectral embedding for data clustering, which linearly combined different graph Laplacians through a set of optimal combination coefficients. Xie \emph{et al.} \cite{Xie2011} proposed a multiview stochastic neighbor embedding for data visualization, which integrated multiple features into a unified representation by learning the combination coefficients of different views based on the robust stochastic gradient descent method.

\section{mHR: Multiview Hessian Regularization}
In multiview semi-supervised learning (mSSL), we are given $l$ labeled examples $\mathcal{L}=\{(x_i^1,x_i^2,...,x_i^{N_v},y_i)\}_{i=1}^l$ and $u$ unlabeled examples $\mathcal{U}=\{x_i^1,x_i^2,...,x_i^{N_v}\}_{i=l+1}^{l+u}$, where $N_v$ is the number of views, $x_i^k\in \mathcal{X}^k$ for $k\in\{1,2,...,N_v \}$ is the $k^{th}$ view feature vector of the $i^{th}$ example (in the following section of this paper, we use $x_i=\{x_i^1,x_i^2,...,x_i^{N_v}\}$ to denote the $i^{th}$ example,$x^k$ denote the $k^{th}$ view feature), $y_i\in \{\pm 1\}$ is the label of  $x_i$. Labeled examples are $(x,y)\in\mathcal{L}$ pairs drawn from a probability $\mathcal{P}$, and unlabeled examples are simply $x\in \mathcal{X}$ drawn according to the marginal distribution $\mathcal{P_X}$ of $\mathcal{P}$, in which $\mathcal{P_X}$ is a compact manifold $\mathcal{M}$. That means the conditional distribution $\mathcal{P}(y|x)$ varies smoothly along the geodesics in the intrinsic geometry of $\mathcal{M}$. And typically $l\ll u$. As usual, the goal is to predict the labels of unseen examples.

By incorporating an additional regularization term to control the complexity of the function along the manifold $\mathcal{M}$, the mSSL problem, similar to SSL, can be written as the following optimization problem
\begin{equation}
\label{eq1}
\min_{f\in \mathcal{H}_K }\frac{1}{l}\sum_{i=1}^l {\psi (f,x_i,y_i)}+\mathcal{G}(\| f \|) ,
\end {equation}
where  $\mathcal{G}(\| f \|)=\gamma_A \| f \|_K^2+\gamma_I\|f\|_K^2$, $\|f\|_K^2$ is the classifier complexity penalty term in an appropriate reproducing kernel Hilbert space (RKHS) $\mathcal{H}_K$, $\|f\|_I^2$ penalizes  $f$ along the compact manifold $\mathcal{M}$, $\psi$ is a general loss function, and parameters  $\gamma_A$ and $\gamma_I$ balance the loss function and regularizations $\|f\|_K^2$ and $\|f\|_I^2$ respectively.

It transpires that the regularization term \(\|f\|_I^2\), estimated from the unlabeled examples, can help us explore the geometry of the marginal distribution  \(\mathcal{P_X}\) which is usually unknown in practice. Although there are different choices for \(\|f\|_I^2\), Laplacian regularization (LR) \cite{Belkin2001,Belkin2006} has received intensive attention. In mSSL, it is important to precisely explore the local geometry of \(\mathcal{M}\), because of the underlying assumption that close examples \(x_i\) and \(x_j\) indicate similar conditional distributions \(\mathcal{P}(y_i |x_i )\) and \(\mathcal{P}(y_j |x_j )\). However, LR biases the classification function \(f\) towards a constant function \cite{Kim}, even though it can only handle examples represented by a single view feature.

In this paper, we introduce the multiview Hessian regularization (mHR) to mSSL-based image annotation. The proposed mHR contains \(N_v\) Hessian regularizations  \(H^k (f)\) for \(k \in {1,2,...,N_v }\), each of which is the matrix of the second order derivative of  \(f\) with respect to the \(k^{th}\) view feature \(x^k\). It varies along the coordinate system and the coordinate system changes along the manifold. Thus, the Frobenius norm of  \(H^k (f)\) is used as a regularizer for encoding an intriguing generalization of the thin-plate splines, because it is invariant along the manifold. We summary the computation of \(H^k (f)\) for the \(k^{th}\) view in the following 4 steps.

\begin{itemize}
  \item For each example \(x_i^k\), we find its \(k\)-nearest neighbors and denote the collection of neighbors as \(\mathcal{N}_i^k\). Form a \(k\times(l+u)\) matrix  \(X_i^k\) whose rows consist of the centralized examples \(x_j^k-x_i^k\) for all \(j\in \mathcal{N}_i^k\). This is different from Hessian Eigenmaps \cite{Donoho2003} that takes the mean of \(x_i^k\) and its \(k\)-nearest neighbors. We choose a different centralization method according to \cite{Kim} to obtain a robust estimation.
  \item Estimate the tangent space by performing a singular value decomposition of  \(X_i^k=UDV^T\), the first \(m\) columns of  \(U\) with the largest \(m\) eigenvalues give the tangent coordinates of examples in \(\mathcal{N}_i^k\).
  \item Perform the Gram-Schmidt orthonormalization process on the matrix  \(M_i^k=[1\ U_1\ ...\ U_m\  U_{11}\  U_{12}\ ...\ U_{mm} ]\) which consists of the following columns: the first column is a vector of ones, and then the first  \(m\) columns of  \(U\); the last \(m(m+1)/2\) columns consist of the various cross-products and squares of those \(m\) columns yielding a matrix \(\hat{M}_i^k\), and then taking the last \(m(m+1)/2\) columns of \(\hat{M}_i^k\) as \(H_i\). The Frobenius norm of the Hessian of  \(f\) on example \(x_a^k\), \(x_b^k (a,b\in\mathcal{N}_i^k)\) at example \(x_i^k\) is \(H_{iab}=\{(H_i )^T H_i\}_{ab}\).
  \item Summing up all the matrices \(H_{iab}\) yields an accumulated matrix \(H^k (f)\) and then the Hessian regularization for the \(k^{th}\) view is given by \(\mathbf{f}^T H^k \mathbf{f}\), \(\mathbf{f}=[f(x_1 ),f(x_2 ),...,f(x_{l+u} )]^T\).
\end{itemize}

In contrast to the LR, mHR enjoys the following advantages: (1) mHR constructs a Hessian regularization (HR) for each view which has a rich nullspace and drives the learned classification function which varies linearly along the manifold; and (2) mHR explores the complementary properties of multiview features. Thus, mHR can better exploit the intrinsic geometry of the marginal distribution \(\mathcal{P_X}\).

\subsection{The General Framework}
In multi-view learning, examples are represented by multiple features. The proposed framework integrates multiple kernel learning and ensemble graph learning.

We first construct a new kernel from kernels defined on each view. Suppose  \(K^k\), \(k=1,...,N_v\) is a valid (symmetric, positive definite) kernel on the \(k^{th}\) view, and then we define the new multiview kernel
\begin{eqnarray}
\label{eq2}
\textbf{\emph{K}}=\sum_{k=1}^{N_v}{\theta^k K^k}, \hspace{1in}\\
s.t. \sum_{k=1}^{N_v}{\theta^k}=1, \theta^k\geq 0, k=1,...,N_v.\nonumber
\end{eqnarray}

Given a set of valid kernels \(\mathcal{K}=\{K^1,K^2,...,K^{N_v }\}\), we denote the convex hull of the set \(\mathcal{A}\) as
\begin{eqnarray}
\mathbf{conv}\mathcal{A}=\{\sum_{k=1}^{N_v}{\theta^k A^k}|\sum_{k=1}^{N_v}{\theta^k}=1, A^k\in \mathcal{A}, \theta^k\geq 0,\nonumber \\
 k=1,...,N_v \}. \nonumber
\end{eqnarray}

Therefore, we have \(\emph{\textbf{K}}\in conv\mathcal{K}\). In the Appendix, we show that \(\textbf{\emph{K}}\) is a valid (symmetric, positive definite) kernel. We then have the regularization term
\begin{eqnarray}
\|f\|_K^2=\mathbf{f}^T K \mathbf{f}=\mathbf{f}^T (\sum_{k=1}^{N_v}{\theta^k K^k}) \mathbf{f}=\sum_{k=1}^{N_v} {\theta^k {\|f\|_K^2}_{(k)}}.\nonumber
\end{eqnarray}

Subsequently, we approximate the intrinsic geometry of a manifold using the convex hull of manifold candidates on each view. Suppose \(H^j\) is the Hessian of the \(j^{th}\) view, we denote
\begin{eqnarray}
\label{eq3}
\textbf{\emph{H}}=\sum_{j=1}^{N_v}{\beta^j H^j}, \hspace{1in}\\
s.t. \sum_{j=1}^{N_v}{\beta^j} =1,\beta^j \geq 0,j=1,...,N_v.\nonumber
\end{eqnarray}

If we define a set of Hessians \(\mathcal{H}=\{H^1,H^2,...,H^{N_v }\}\), we have \(\textbf{\emph{H}}\in conv\mathcal{H}\) and mHR is defined by
\begin{eqnarray}
\|f\|_I^2=\mathbf{f}^T \textbf{\emph{H}}\mathbf{f}=\mathbf{f}^T (\sum_{j=1}^{N_v}{\beta^j H^j})\mathbf{f}=\sum_{j=1}^{N_v}{\beta^j {\|f\|_I^2}_{(j)}}.\nonumber
\end{eqnarray}

Therefore, we can obtain the mHR framework for multiview learning
\begin{eqnarray}
\label{eq4}
\min_{f \in \mathcal{H}_K, \theta \in \mathbf{R}^{N_v}, \beta \in \mathbf{R}^{N_v}}{\frac{1}{l} \sum_{i=1}^l {\psi (f,x_i,y_i)} + \gamma_A \sum_{k=1}^{N_v}{\theta^k {\|f\|_K^2}_{(k)}}} \nonumber \\
{ + \gamma_I \sum_{j=1}^{N_v}{\beta^j {\|f\|_I^2}_{(j)}} + \gamma_\theta \|\theta\|_2^2 + \gamma_\beta \|\beta\|_2^2 }, \hspace{0.1in}\\
s.t. \sum_{k=1}^{N_v}\theta^k =1,\theta^k \geq 0,k=1,...,N_v, \hspace{0.5in}\nonumber \\
\sum_{j=1}^{N_v}\beta^j =1,\beta^j \geq 0,j=1,...,N_v, \hspace{0.5in}\nonumber
\end{eqnarray}
where the regularization terms \(\|\theta\|_2^2\) and \(\|\beta\|_2^2\) are introduced to avoid the model parameter overfitting to only one view kernel or manifold, and \(\gamma_\theta \in \mathbf{R}^+ \)and \(\gamma_\beta \in \mathbf{R}^+\) are the trade-off parameters to control the contributions of the regularization terms \(\|\theta\|_2^2\) and \(\|\beta\|_2^2\) respectively.

For fixed \(\theta\) and \(\beta\), (\ref{eq4}) degenerates to (\ref{eq1}), with \( \textbf{\emph{K}}=\sum_{k=1}^{N_v}{\theta^k K^k} \)and \(\textbf{\emph{H}}=\sum_{j=1}^{N_v}{\beta^j H^j}\).

On the other hand, for fixed \(f\) and \(\beta\), (\ref{eq4}) can be simplified to:
\begin{eqnarray}
\label{eq5}
\theta^*={\arg\min}_{\theta \in \mathbf{R}^{N_v } }{\sum_{k=1}^{N_v}{\theta^k h^k} +\theta^T \textbf{\emph{B}} \theta},\\
s.t. \sum_{k=1}^{N_v}\theta^k =1,\theta^k \geq 0,k=1,...,N_v),\nonumber \\
 \textbf{\emph{B}}=\textbf{\emph{K}}^T \textbf{\emph{H}} \textbf{\emph{K}}.\hspace{1.3in}\nonumber
\end{eqnarray}

And for fixed \(f\) and \(\theta\), (\ref{eq4}) can be simplified to:
\begin{eqnarray}
\label{eq6}
\theta^*={\arg\min}_{\beta \in \mathbf{R}^{N_v } }{\sum_{k=1}^{N_v}{\beta^k h^k} + \gamma_\beta \|\beta\|_2^2},\\
s.t. \sum_{k=1}^{N_v}\beta^k =1,\beta^k \geq 0,k=1,...,N_v.\nonumber
\end{eqnarray}

The solution of (\ref{eq5}) and (\ref{eq6}) can be viewed as the learning of the optimal linear combination of the kernels or Hessians over different views.

\subsection{Representer Theorem and Convergence Analysis}

This section shows the representer theorem of mHR and the convergence analysis for the alternating optimization of (\ref{eq4}). Detailed proofs are given in the appendix. We first show the following lemmas which are essential for the representer theorem.

\emph{Lemma 1:} If  \(\mathcal{G}(\|f\| )\) is a strictly monotonically increasing real-valued function with respect to  \(\|f\|\), the minimizer of the optimization problem (\ref{eq1}) admits an expansion
\begin{eqnarray}
f^*=\sum_{i=1}^{l+u}{\alpha_i K(x_i,x)} \nonumber
\end{eqnarray}
in terms of the labeled and unlabeled examples.

\emph{Lemma 2:} \(\textbf{\emph{K}}\in conv\mathcal{K}\) is a valid kernel.

\emph{Lemma 3:} \(\textbf{\emph{H}}\in conv\mathcal{H}\) is semi-definite positive.

\textbf{Theorem 1.} The minimization of (\ref{eq4}) w.r.t. \(f\) with fixed \(\theta\) and \(\beta\), exits and has the representation
\begin{equation}
\label{eq7}
f^*=\sum_{i=1}^{l+u}{\alpha_i \textbf{\emph{K}}(x_i,x)}=\sum_{i=1}^{l+u}{\alpha_i \sum_{k=1}^{N_v}{\theta^k K^k (x_i^k,x^k)}},
\end{equation}
which is an expansion in terms of the labeled and unlabeled example.

The representer theorem shows the solution of (\ref{eq4}) exists and has the general form of (\ref{eq7}) given fixed \(\beta\) and  \(\theta\). The purpose of mHR is to learn the classifier  \(f\), and the combination coefficients  \(\beta\) and  \(\theta\). In this paper, we use the alternating optimization \cite{Bezdek2003} to iteratively solve (\ref{eq4}). First, fix \(\theta\) and \(\beta\) to optimize  \(\alpha\). Then, fix \(\alpha\) and \(\beta\) to optimize  \(\theta\). Finally, fix \(\alpha\) and \(\theta\) to optimize  \(\beta\). We conduct the above three steps iteratively until convergence. The above alternating iteration process is convergent, and the convergence theorem is given below.

\textbf{Theorem 2.} Given a convex loss function  \(\psi\), the alternating optimization for solving (\ref{eq4}) produces a monotonically decreasing sequence that converges to a local minimum.

The proof of Theorem 2 shows that for a convex loss function, (\ref{eq4}) is convex w.r.t. \(f\) for fixed \(\theta\) and \(\beta\), and w.r.t \(\theta\) (or \(\beta\)) for fixed \(f\) and \(\beta\) (or \(\theta\)). However, (\ref{eq4}) is not convex for \((f,\theta,\beta)\) jointly. Fortunately, we can initialize \(\theta^k=1/N_v\)  and \(\beta^k=1/N_v\)  for all \(k=1,...,N_v\). This initialization empirically results in a satisfied solution of (\ref{eq4}).

\section{Example Algorithms}

Generally, \(\psi(f,x_i,y_i )\) can be any loss function and mHR can be applied to general purpose mSSL-based applications. In this section, we show the implementations of mHR through kernel least squares (KLS) and SVM.

\subsection{mHR support vector machines (mHR-SVM)}

SVM minimizes the hinge loss, \emph{i.e.} \(\psi(f,x_i,y_i )={(1-y_i f(x_i))}_+=\max{(0,1-y_i f(x_i))}\). By introducing mHR to SVM, we can obtain mHR-SVM
\begin{eqnarray}
\label{eq8}
\min_{f \in \mathcal{H}_K,\theta \in \mathbf{R}^{N_v},\beta \in \mathbf{R}^{N_v} }{\gamma_A \sum_{k=1}^{N_v}{\theta^k {\|f\|_K^2}_{(k)}} +\gamma_I \sum_{j=1}^{N_v}{\beta^j {\|f\|_I^2}_{(j)}}} \nonumber \\
{+ \frac{1}{l} \sum_{i=1}^l{{(1-y_i f(x_i))}_+} + \gamma_\theta \|\theta\|_2^2 + \gamma_\beta \|\beta\|_2^2},\\
s.t. \sum_{k=1}^{N_v}\theta^k =1,\theta^k \geq 0,k=1,...,N_v, \hspace{0.5in}\nonumber \\
\sum_{j=1}^{N_v}\beta^j =1,\beta^j \geq 0,j=1,...,N_v.\hspace{0.5in}\nonumber
\end{eqnarray}

Given fixed \(\beta\) and \(\theta\), (\ref{eq8}) can be expressed as following by substituting (\ref{eq7}) into (\ref{eq8})
\begin{eqnarray}
\label{eq9}
\min_{\alpha\in\mathbf{R}^{l+u},\theta\in\mathbf{R}^{N_v },\beta\in\mathbf{R}^{N_v}} {\gamma_A \alpha^T \textbf{\emph{K}} \alpha + \gamma_I \alpha^T \textbf{\emph{KHK}} \alpha   } \nonumber \\
{+\frac{1}{l} \sum_{i=1}^l{{(1-y_i \textbf{\emph{K}}(x_i,x)\alpha)}_+} + \gamma_\theta \|\theta\|_2^2 + \gamma_\beta \|\beta\|_2^2}, \\
s.t. \sum_{k=1}^{N_v}{\theta^k} =1,\theta^k \geq 0,k=1,...,N_v, \nonumber\\
 \sum_{j=1}^{N_v}{\beta^j} =1,\beta^j \geq 0,j=1,...,N_v, \nonumber
\end{eqnarray}
where \(\textbf{\emph{K}}=\sum_{k=1}^{N_v}{\theta^k K^k}\), \(\textbf{\emph{H}}=\sum_{j=1}^{N_v}{\beta^j H^j}\), and \({(1-y_i \textbf{\emph{K}}(x_i,x)\alpha)}_+=\max{(0,1-y_i \textbf{\emph{K}}(x_i,x)\alpha)}\) is the hinge loss function.

Given fixed \(\theta\) and \(\beta\), (\ref{eq9}) can be rewritten as
\begin{equation}
\label{eq10}
\min_{\alpha \in \mathbf{R}^{l+u}}{F(\alpha)=R(\alpha)+\psi(\alpha)},
\end{equation}
where \(R(\alpha)=\gamma_A \alpha^T \textbf{\emph{K}} \alpha +\gamma_I \alpha^T \textbf{\emph{KHK}} \alpha + \gamma_\theta  \|\theta\|_2^2 + \gamma_\beta \|\beta\|_2^2 \) and  \(\psi(\alpha)=\frac{1}{l} \sum_{i=1}^l{{(1-y_i \textbf{\emph{K}}(x_i,x)\alpha)}_+} \).

The loss function part \(\psi(\alpha)\) is non-differentiable. Hence, we firstly smooth the hinge loss and then use Nesterov's optimal gradient method \cite{Nesterov2005a} to solve (\ref{eq10}). In the \(t^{th}\) iteration round, two auxiliary optimizations are introduced to compute the solution. Suppose \(F_\mu(\alpha)\) is a smooth function of  \(F(\alpha)\),  \(L_\mu\) is the Lipschitz constant of \(F_\mu (\alpha)\), \(\alpha^{(t)}\) is the solution at the \(t^{th}\) iteration, the two auxiliary optimizations are given by
\begin{eqnarray}
\min_{\mathrm{y} \in \mathbf{R}^{l+u} }{<\nabla F_\mu(\alpha^{(t)}),\mathrm{y}-\alpha^{(t)}> + \frac{L_\mu}{2} \|\mathrm{y}-\alpha^{(t)}\|_2^2}, \nonumber
\end{eqnarray}
and
\begin{eqnarray}
\min_{z \in \mathbf{R}^{l+u}}{ \sum_{i=1}^t{\frac{i+1}{2} [F_\mu (\alpha^{(i)} ) + < \nabla F_\mu (\alpha^{(i)}),z-\alpha^{(i)}>]}} \nonumber \\
{+ \frac{L_\mu}{2} \|z-\hat{\alpha}\|_2^2 }, \nonumber
\end{eqnarray}
where \(\hat{\alpha}\) is a guess solution of \(\alpha\).

The solutions of the two optimizations are
\begin{eqnarray}
\label{eq11}
\mathrm{y}^{{t}}=\alpha^{(t)}-\frac{1}{L_\mu} \nabla F_\mu(\alpha^{(t)}),
\end{eqnarray}
\begin{eqnarray}
\label{eq12}
z^{(t)}=\hat{\alpha}-\frac{1}{L_\mu} \sum_{i=1}^t{\frac{i+1}{2} \nabla F_\mu (\alpha^{(i)})}.
\end{eqnarray}

By using the weighted sum of \(\mathrm{y}^{(t)}\) and \(z^{(t)}\), we obtain the solution of (\ref{eq9}) after the \(t^{th}\) iteration round,
\begin{eqnarray}
\label{eq13}
\alpha^{(t+1)}=\frac{2}{t+3} z^{(t)} + \frac{t+1}{t+3} \mathrm{y}^{(t)}.
\end{eqnarray}
According to \cite{Nesterov2005a}, the hinge loss can be smoothed by subtracting a strongly convex function from its saddle point function. The smoothed hinge loss can then be written as
\begin{eqnarray}
\label{eq14}
\psi_\mu=\max_{\mathbf{u} \in \mathcal{Q}}{\mathbf{u}_i (1-y_i \textbf{\emph{K}}(x_i,x)\alpha) - \frac{\mu}{2} \|\textbf{\emph{K}}(x_i,x)\|_\infty \mathbf{u}_i^2}, \\
\mathcal{Q}=\{\mathbf{u}: 0 \leq \mathbf{u}_i \leq 1, \mathbf{u} \in \mathbf{R}^l \}, \nonumber
\end{eqnarray}
where \(\mu\) is the smooth parameter. To solve (\ref{eq14}), \(\mathbf{u}_i\) can be computed and projected on \(\mathcal{Q}\) by using
\begin{eqnarray}
\label{eq15}
\mathbf{u}_i=median\{0,1,\frac{1-y_i \textbf{\emph{K}}(x_i,x)\alpha}{\mu\|\textbf{\emph{K}}(x_i,x)\|_\infty }\}.
\end{eqnarray}

Then the gradient \(\nabla F_\mu (\alpha^t )\) is
\begin{eqnarray}
\label{eq16}
\nabla F_\mu(\alpha^t)=2(\gamma_A \textbf{\emph{K}} + \gamma_I \textbf{\emph{KHK}}) \alpha - \frac{1}{l}{(Y\textbf{\emph{K}}(x_i,x))}^T \mathbf{u},
\end{eqnarray}
where \(Y=diag(y)\).

The Lipschitz constant of  \(\psi(\alpha)\) is

\begin{eqnarray}
L^\psi=\frac{1}{\mu} \max_i{\frac{\|\textbf{\emph{K}}(x_i,x)^T \textbf{\emph{K}}(x_i,x)\|_2}{\|\textbf{\emph{K}}(x_i,x)\|_\infty}}. \nonumber
\end{eqnarray}

Then the Lipschitz constant of \(F_\mu(\alpha)\) is given by
\begin{eqnarray}
\label{eq17}
F_\mu = L^R + L^\psi = \|2(\gamma_A \textbf{\emph{K}} + \gamma_I \textbf{\emph{KHK}})\|_2 \nonumber \\
+ \frac{1}{\mu} \max_i{\frac{\|\textbf{\emph{K}}(x_i,x)^T \textbf{\emph{K}}(x_i,x)\|_2}{\|\textbf{\emph{K}}(x_i,x)\|_\infty}}.
\end{eqnarray}

After substituting (\ref{eq17}) into (\ref{eq11}), (\ref{eq12}), we obtain \(\alpha\) according to (\ref{eq13}).

The procedure for optimizing \(\theta\) given fixed \(\alpha\) and \(\beta\) is the same as that for optimizing \(\beta\) given fixed \(\alpha\) and \(\theta\). The convergence analysis in Theorem 2 ensures the alternating optimization obtains a local optimal solution. In this paper, we initialize \(\theta^k=1/N_v\)  and \(\beta^k=1/N_v\)  for all \(k=1,...,N_v\), and update them by using the coordinate descent method.

\begin{figure*}[!t]
\centering
\includegraphics[width=5in]{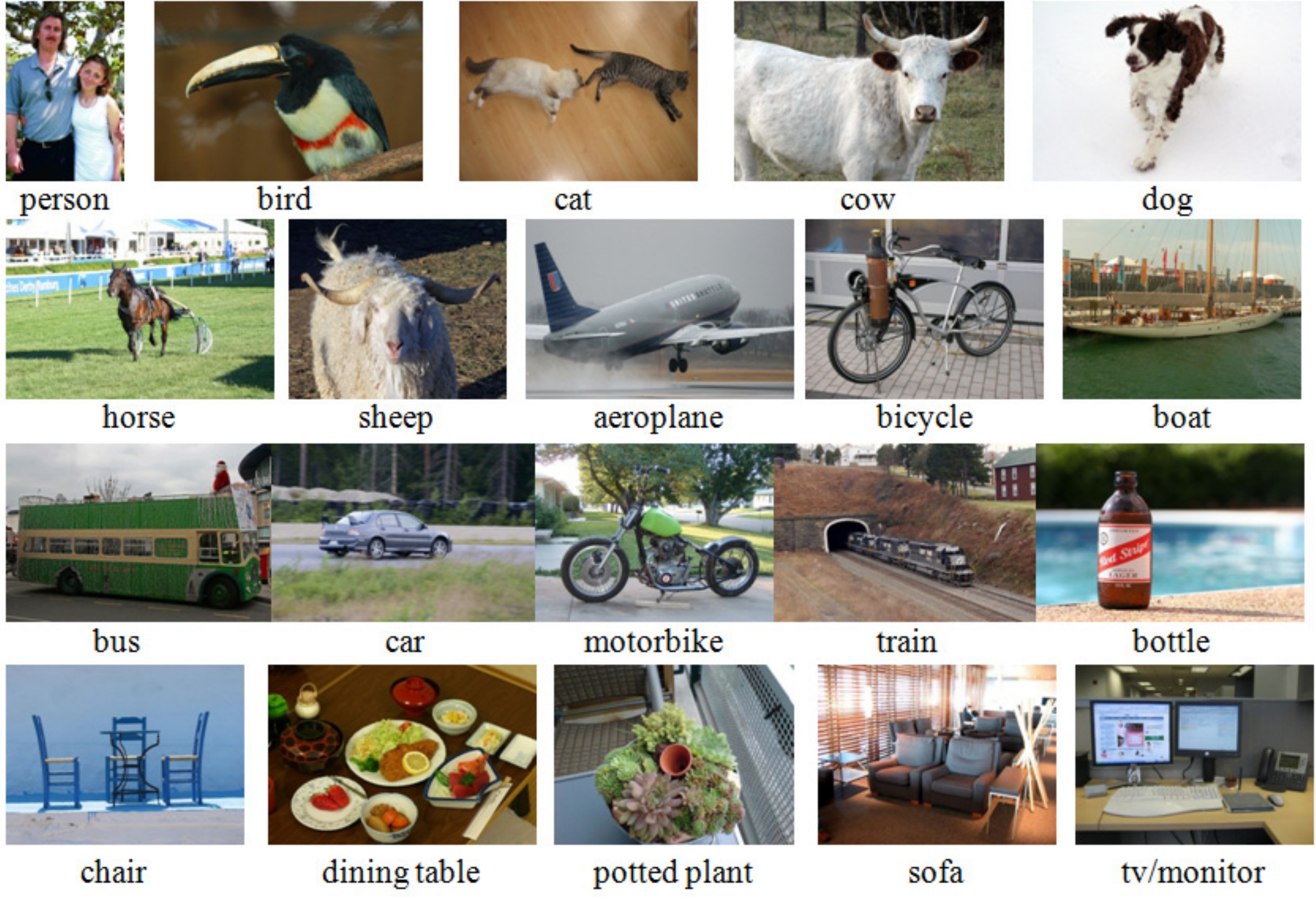}
\caption{Example images of PASCAL VOC'07 including person, bird, cat, cow, dog, horse, sheep, aeroplane, bicycle, boat, bus, car, motorbike, train, bottle, chair, dining table, potted plant, sofa, tv/monitor.}
\label{fig1}
\end{figure*}
\begin{table*}[!t]
\caption{List of algorithms}
\label{Table1}
\centering
\begin{tabular}{|l|l|l|}
\hline
abbreviation & method & feature/view\\
\hline
\hline
SVM & support vector machine & different visual feature/single view\\
\hline
LapSVM &	Laplacian regularized SVM &	different visual feature/single view\\
\hline
HesSVM &	Hessian regularized SVM	& different visual feature /single view\\
\hline
ConSVM &	support vector machine &	concatenation of 15 different visual features /multiview\\
\hline
LapCSVM &	Laplacian regularized SVM &	concatenation of 15 different visual features /multiview\\
\hline
HesCSVM &	Hessian regularized SVM &	concatenation of 15 different visual features /multiview\\
\hline
AveSVM &	support vector machine	& average of 15 different kernels /multiview\\
\hline
LapASVM &	Laplacian regularized SVM	& average of 15 different kernels /multiview\\
\hline
HesASVM	& Hessian regularized SVM	& average of 15 different kernels /multiview\\
\hline
mHesSVM	& multi-view Hessian regularized SVM	& multiple kernels /multiview\\
\hline
\hline
KLS &	Kernel least squares	& different visual feature /single view\\
\hline
LapLS &	Laplacian regularized least squares	& different visual feature /single view\\
\hline
HesLS & Hessian regularized least squares	& different visual feature /single view\\
\hline
ConLS &	Kernel least squares 	& concatenation of 15 different visual features /multiview\\
\hline
LapCLS &	Laplacian regularized least squares	& concatenation of 15 different visual features /multiview\\
\hline
HesCLS &	Hessian regularized least squares	& concatenation of 15 different visual features /multiview\\
\hline
AveLS &	Kernel least squares 	& average of 15 different kernels /multiview\\
\hline
LapALS &	Laplacian regularized least squares	& average of 15 different kernels /multiview\\
\hline
HesALS &	Hessian regularized least squares	& average of 15 different kernels /multiview\\
\hline
mHesLS &	multi-view Hessian regularized LS	& multiple kernels /multiview\\
\hline
\end{tabular}
\end{table*}

\subsection{mHR kernel least squares (mHR-KLS)}

The loss in regularized kernel least squares is defined by the squared loss, \emph{i.e.} \(\psi(f,x_i,y_i )={(y_i-f(x_i))}^2\). By introducing mHR to regularized KLS, we have mHR-KLS
\begin{eqnarray}
\label{eq18}
\min_{f \in \mathcal{H}_K,\theta \in \mathbf{R}^{N_v},\beta \in \mathbf{R}^{N_v}}{ \gamma_A \sum_{k=1}^{N_v}{\theta^k {\|f\|_K^2}_{(k)}} + \gamma_I \sum_{j=1}^{N_v}{\beta^j {\|f\|_I^2}_{(j)}} } \nonumber \\
{+ \frac{1}{l} \sum_{i=1}^l{{(y_i - f(x_i))}^2} + \gamma_\theta \|\theta\|_2^2 + \gamma_\beta \|\beta\|_2^2}.
\end{eqnarray}

According to Theorem 1, given fixed \(\beta\) and \(\theta\), we substitute (\ref{eq7}) into (\ref{eq18}) and obtain
\begin{eqnarray}
\label{eq19}
\min_{\alpha \in \mathbf{R}^{l+u},\theta \in \mathbf{R}^{N_v },\beta \in \mathbf{R}^{N_v}} { \gamma_A \alpha^T \textbf{\emph{K}} \alpha  + \gamma_I \alpha^T \textbf{\emph{KHK}} \alpha}\nonumber \\
{+\frac{1}{l} \sum_{i=1}^l{{(Y-J\textbf{\emph{K}}\alpha)}^T (Y-J\textbf{\emph{K}}\alpha)} + \gamma_\theta \|\theta\|_2^2 + \gamma_\beta \|\beta\|_2^2}, \\
s.t. \sum_{k=1}^{N_v}{\theta^k} =1,\theta^k \geq 0,k=1,...,N_v,  \nonumber\\
 \sum_{j=1}^{N_v}{\beta^j} =1,\beta^j \geq 0,j=1,...,N_v, \nonumber
\end{eqnarray}
where  \(\textbf{\emph{K}}=\sum_{k=1}^{N_v}{\theta^k K^k}\), \(\textbf{\emph{H}}=\sum_{j=1}^{N_v}{\beta^j H^j}\),\ \(Y=[y_1,y_2,...,y_l,0,...,0] \in \mathbf{R}^{l+u}\) is an \((l+u)\)-dimensional vector and \(J \in \mathbf{R}^{l+u} \times \mathbf{R}^{l+u}\) is a diagonal matrix with the first \(l\) diagonal elements as 1 and the rest 0.

The solution of (\ref{eq19}) can be obtained by setting the derivatives of its object function w.r.t \(\alpha\) to zero. Then we have
\begin{eqnarray}
\label{eq20}
\alpha^* = {(J\textbf{\emph{K}} + \gamma_A l I_{l+u} + \gamma_I l \textbf{\emph{HK}})}^{-1} Y,
\end{eqnarray}
where \(I_{l+u}\in \mathbf{R}^{l+u} \times \mathbf{R}^{l+u}\) is an identity matrix.

The procedure for optimizing \(\theta\) given fixed \(\alpha\) and \(\beta\) and optimizing \(\beta\) given fixed \(\alpha\) and \(\theta\) is similar to that used in mHR-SVM. The convergence analysis in Theorem 2 ensures the alternating optimization obtains a local optimal solution. In this paper, we initialize \(\theta^k=1/N_v\)  and \(\beta^k=1/N_v\)  for all \(k=1,...,N_v\), and update them by using the coordinate descent method.

\begin{figure*}[!t]
\centering
\includegraphics[width=7in]{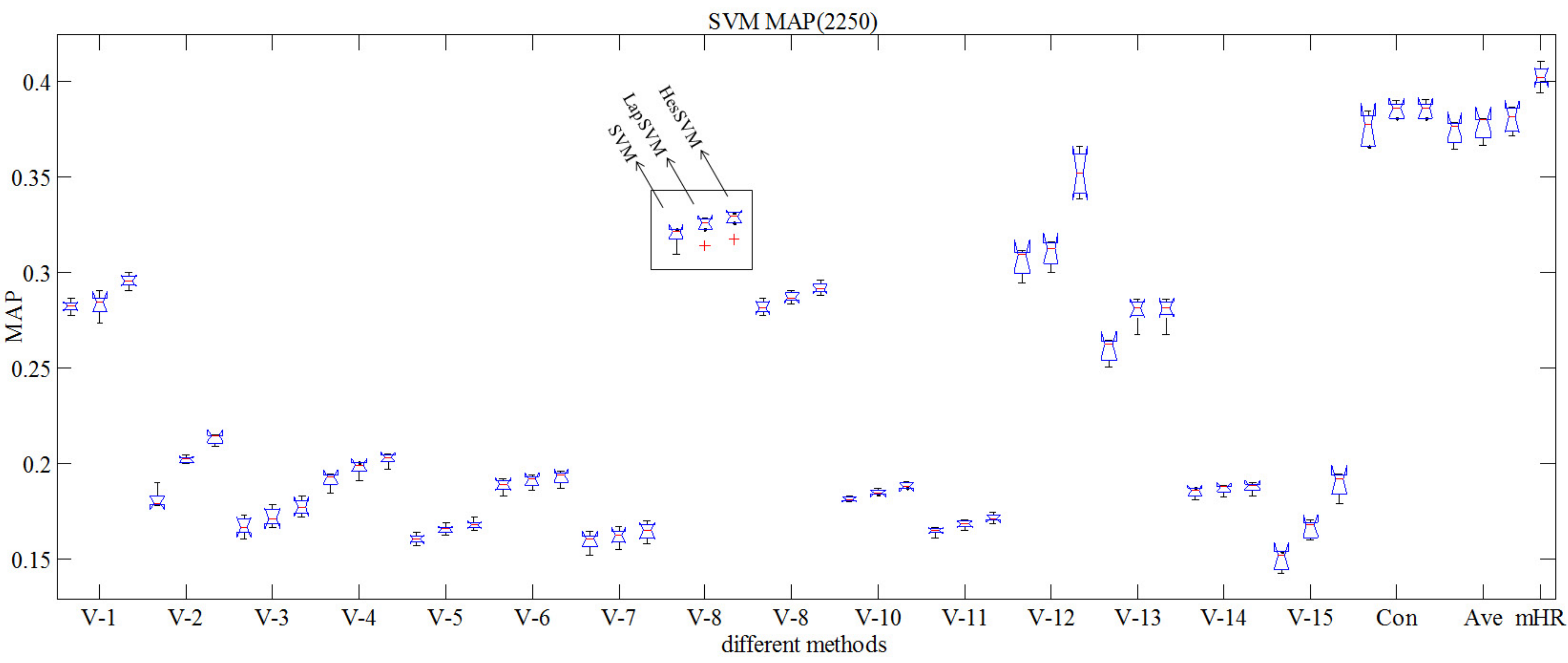}
\caption{The mAP of different SVM methods on 2250 labeled images. For each feature/view, the methods from left to right are SVM, LapSVM and HesSVM.}
\label{fig2}
\end{figure*}

\section{Experiments}

To evaluate the effectiveness of the proposed mHR, we apply mHR-SVM and mHR-KLS to image annotation \cite{Caoliangliang2010} and conduct experiments on the PASCAL VOC'07 dataset \cite{EveringhamL2007}. This dataset contains 9,963 images of 20 visual object classes. Figure 1 shows 20 example images sampled from these 20 classes.

In our experiments, 15 visual features provided by Guillaumin \emph{et al.} \cite{Guillaumin2010} are used, including GIST feature, 2 RGB features, 2 Lab features, 2 HSV features, 2 Hue features, 2 SIFT features, 2 Harris features, and 2 Harris+SIFT features.

We use the standard training/test partition according to \cite{EveringhamL2007}, in which the training set contains 5,011 images and the test set contains 4,952 images. To choose suitable model parameters, we divide the training set into two subsets; one contains 4,500 images for model training and one contains 511 images for model parameter tuning. We randomly divide the training set 10 times to examine the robustness of different learning models. In the semi-supervised learning experiments, in particular, we assign 10\%, 20\%, 30\%, 50\%, 70\% and 90\% as labeled data and the rest as unlabeled data. All parameters are tuned in the case of 10\% labeled data and 90\% unlabeled data. Particularly, parameters \(\gamma_A\), \(\gamma_I\), \(\gamma_\theta\) and \(\gamma_\beta\) are tuned from the candidate set \(\{10^e |e=-10,-9,...,9,10\}\) on validation set in the case of 10\% labeled data and 90\% unlabeled data. And the parameter \(k\) which is the number of the neighbors in \(k\)-nearest neighbors in computing Hessian and graph Laplacian is fixed to 100 for all experiments.

We compare the proposed mHR-SVM with SVM and Laplacian regularized SVM (LapSVM), and compare mHR-KLS with KLS and Laplacian regularized KLS (LapKLS). By replacing the Laplacian matrix in LapSVM and LapKLS with the Hessian matrix, we obtain Hessian regularized SVM (HesSVM) and Hessain regularized KLS (HesKLS), respectively. HesSVM and HesKLS are important to validate the effectiveness of Hessian regularization in SSL. To comprehensively examine the effectiveness of mHR, we also compare mHR-SVM (mHR-KLS) with the feature concatenation method (by concatenating 15 different visual features into a long feature vector) and average kernel method (by taking the average of 15 different kernels, each of which is obtained from a particular visual feature) used in SVM (KLS), LapSVM (LapKLS), and HesSVM (HesKLS). In summary, we have 18 baseline algorithms listed in Table 1.

We use the average precision (AP) \cite{Wangmeng2009} for each class and mean average precision (mAP) of all classes as measure criteria. In our experiments, the AP and mAP are computed using PASCAL VOC method \cite{EveringhamL2007}, \emph{i.e.}
\begin{eqnarray}
AP=\frac{1}{11}\sum_t{[\max_{p(k)\geq t}{p(k)}]},t \in \{0, 0.1, 0.2, ..., 1.0\},\nonumber
\end{eqnarray}
and
\begin{eqnarray}
mAP=\frac{\sum_{i=1}^\#{AP_i}}{\#\{visual\ object\ classes\}}, \nonumber
\end{eqnarray}
where \(p(k)\) is the precision at the cut-off of the rank index of positive sample \(k\). And to demonstrate the robustness of the algorithms, we report the results (AP and mAP) using the error bars which can show the confidence intervals of data and notched box plot in which five values from a set of data are conventionally used (the extremes, the upper and lower hinges, and the median) and the notches surrounding the medians provide a measure of the rough significance of differences between the values.

\subsection{Effectiveness of HR}

To evaluate the effectiveness of HR, we construct HesSVM (HesLS), LapSVM (LapLS) and SVM (KLS) over 20 visual object classes on different 15 features respectively. We also construct mHesSVM (mHesLS) and the corresponding feature concatenation method and average kernel method over all visual object classes.

Figure 2 is a notched box plot of the mAP of different SVM methods on 2250 labeled images. From Figure 2, we can see that HR performs better than LR in most cases. Figure 2 also shows that the multiview methods are significantly better than single view methods and mHR outperforms other multiview methods.

\begin{figure*}[!t]
\centering
\includegraphics[width=7.1in]{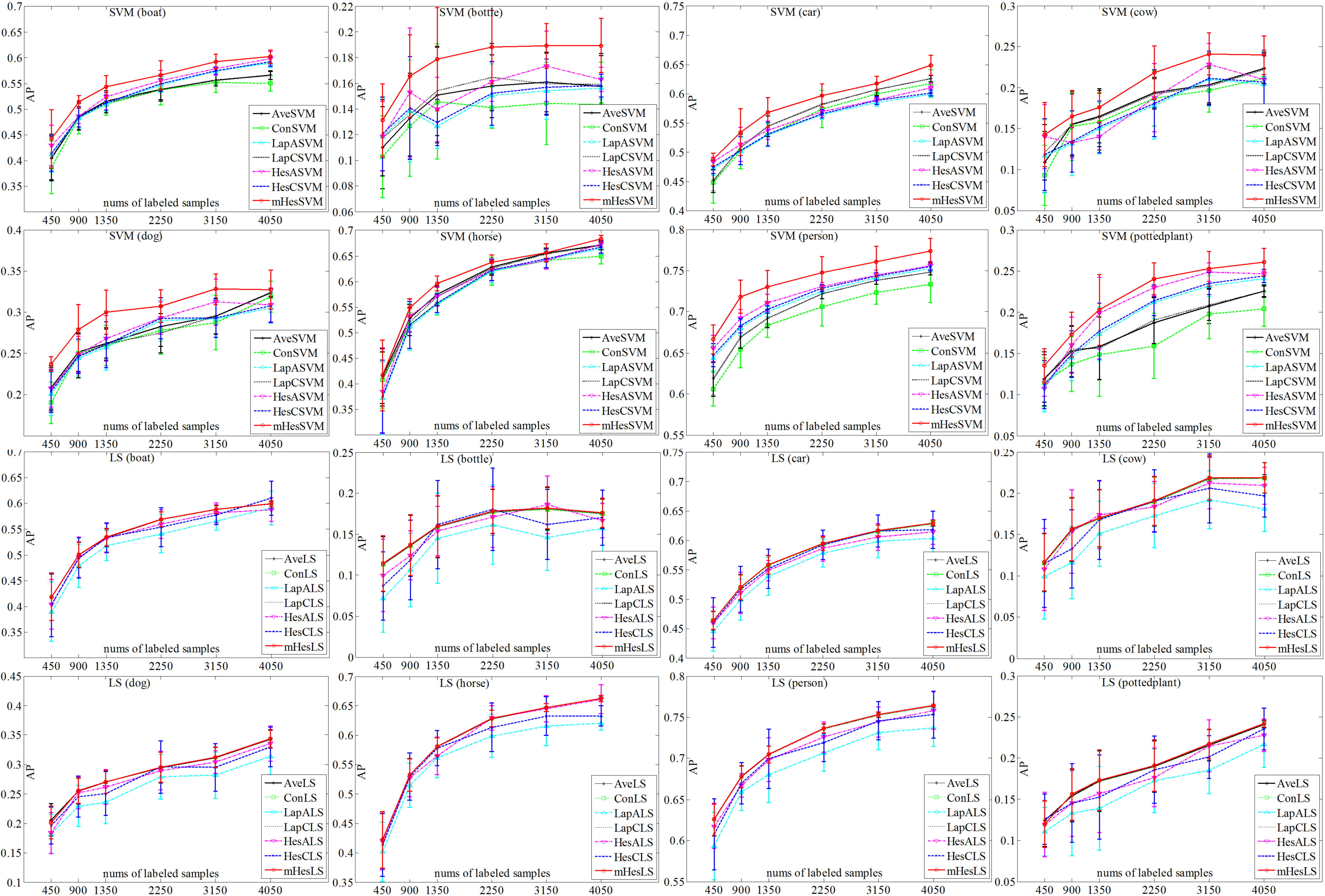}
\caption{The AP of different multiview methods on some classes including boat, bottle, car, cow, dog, horse, person and potted plant. The upper 8 subfigures are SVM methods, and the lower 8 are LS methods.}
\label{fig3}
\end{figure*}

\subsection{Performance of mHR}

To further evaluate the performance of mHR, we compare the AP of mHR with the feature concatenation method and average kernel method over each visual object class. We also compare the mAP of different methods over all classes.

Figure 3 is the AP of different multiview methods on selected visual object classes. Each subfigure corresponds to one visual object class of the dataset. The x-coordinate is the number of labeled images. From Figure 3, we can see that mHR significantly boosts performance, especially when the number of labeled images is small.

Figure 4 is the mAP boxplot of different multiview methods. The subfigures correspond to the performance on different numbers (450, 900, 1350, 2250, 3150 and 4050) of labeled images. From Figure 4, we can see that the average kernel method performs better than the feature concatenation method in SVM implementation. However, the feature concatenation method performs better than the average kernel method in KLS implementation, and in both SVM and KLS implementations, mHR outperforms the other multiview methods.

\begin{figure*}[!t]
\centering
\includegraphics[width=7in]{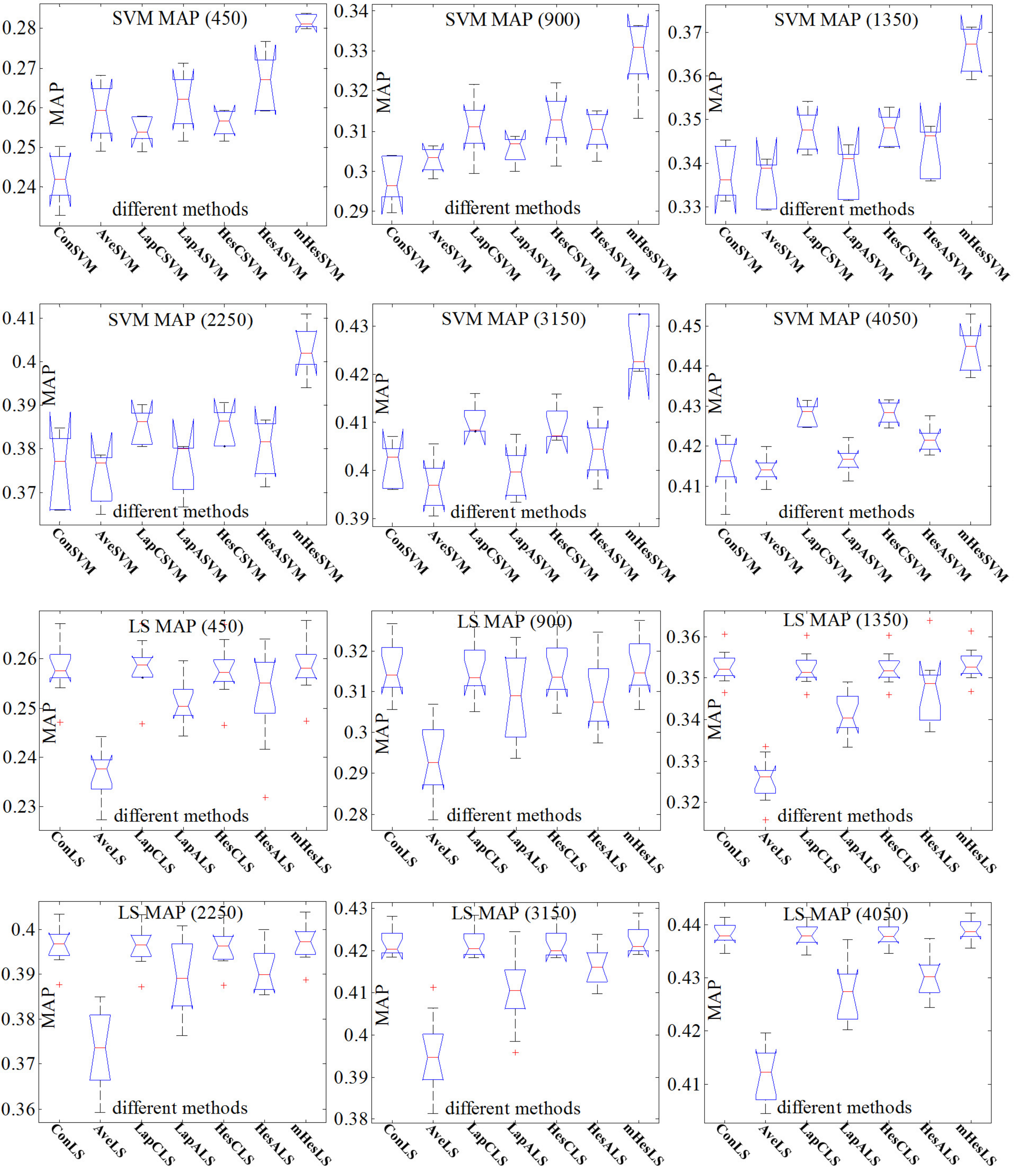}
\caption{The mAP of different multiview methods. The upper 6 subfigures are SVM methods, and the lower 6 are LS methods.}
\label{fig4}
\end{figure*}

\section{Concluding Remarks}
Manifold regularization-based semi-supervised learning algorithms have been successfully applied to image annotation. However, most of the existing methods are based on Laplacian regularization which suffers from the lack of extrapolating power, particularly when the number of labeled examples is small. In addition, conventional methods are often designed to cover single views, which is not applicable to the practical multiview applications. Therefore, we present multiview Hessian regularization (mHR) to tackle the above two problems for image annotation. The proposed mHR can naturally combine both multiple kernels and Hessian regularizations obtained from different views to boost learning performance. With the help of Hessian regularization and the multi-view feature, mHR can steer the learned function which varies linearly along the data manifold and can competently explore the complementary information from different view features. We apply mHR to kernel least squares and support vector machines as two implementations for image annotation. Extensive experiments on the PASCAL VOC'07 dataset demonstrate that the proposed mHR significantly outperforms LR-based and other related algorithms.

\appendices
\section{Proof of Lemma 1}
\begin{IEEEproof}
Suppose we project \(f\) onto the subspace  \(S=span{K(x_i,x):1 \leq i \leq l+u} \) spanned by kernels. Any \(f \in \mathcal{H}_K\) can then be represented as  \(f=f_S+f_{S^\perp}\), where \(f_S\) is the component along the subspace and  \(f_{S^\perp}\) is the component perpendicular to the subspace. Then we have  \(\|f\|^2=\|f_S\|^2 + \|f_{S^\perp}\|^2 \geq \|f_S\|^2\). Since  \(\mathcal{G}(\|f\| )\) is a strictly monotonically increasing real-valued function on \(\|f\|\), we have  \(\mathcal{G}(\|f\|^2 )\geq \mathcal{G}(\|f_S\|^2 )\). This implies that  \(\mathcal{G}(\|f\| )\) is minimized if  \(f\)  lies in the subspace. Note the reproducing property of the kernel  \(K\), then \(f(x_i )=<f,K(x_i,x)>=<f_S,K(x_i,x)> + <f_{S^\perp},K(x_i,x)>=f_S (x_i )\). That means the loss function part of (\ref{eq1}) only depends on \(f_S\). Thus, the minimizer of the optimization problem (\ref{eq1}) can be obtained when  \(f\) lies in the subspace \(S\), that is \(f^*=\sum_{i=1}^{l+u}{\alpha_i K(x_i,x)}\). This completes the proof of Lemma 1.
\end{IEEEproof}
\section{Proof of Lemma 2}
\begin{IEEEproof}
Suppose \(K^k=<\phi_k (w^k ),\phi_k (v^k )>:\mathcal{X} \times \mathcal{X} \mapsto \mathbf{R} \) is valid (symmetric, positive definite) for all \(k=1,...,N_v\). Then we have \(\theta^k K^k (w^k,v^k)=<\sqrt{\theta^k} \phi_k (w^k ),\sqrt{\theta^k} \phi_k (v^k )>\). Then
\begin{eqnarray}
&&\textbf{\emph{K}}(w,v)=\sum_{k=1}^{N_v}{\theta^k K^k (w^k,v^k)} \nonumber \\
&&=\sum_{k=1}^{N_v}{<\sqrt{\theta^k} \phi_k (w^k ),\sqrt{\theta^k} \phi_k (v^k )>} \nonumber \\
&&=<\mathbf{W},\mathbf{V}>, \nonumber
\end{eqnarray}
where \(\mathbf{W}=[\sqrt{\theta^1} \phi_1 (w^1 )...\sqrt{\theta^k} \phi_k (w^k )...\sqrt{\theta^{N_v}} \phi_{N_v} (w^{N_v} )]\), \(\mathbf{V}=[\sqrt{\theta^1} \phi_1 (v^1 )...\sqrt{\theta^k} \phi_k (v^k )...\sqrt{\theta^{N_v}} \phi_{N_v} (v^{N_v} )  ]\).

We can see that \( \textbf{\emph{K}}(w,v)\) can be represented as an inner product, and thus it is also a valid kernel.

\end{IEEEproof}

\section{Proof of Lemma 3}
\begin{IEEEproof}
For any \(\beta\) that satisfies the constraints in (\ref{eq3}), \(\textbf{\emph{H}}=\sum_{j=1}^{N_v}{\beta^j H^j}\) is a convex combination of the Hessian energy in the set \(\mathcal{H}\). According to the computation procedure of Hessian, we have \(H^j=\sum_i{H_{iab}^j},\ i=1,2,...,l+u\), where \(H_{iab}^j=\{{(H_i^j )}^T H_i^j\}_{ab}\) is the Frobenius norm of the Hessian of  \(f\) on example \(x_a^j,x_b^j (a,b \in \mathcal{N}_i^j)\)  at example \(x_i^j\). \(H_{iab}^j\) is semi-definite positive, then we have \(H^j\) is semi-definite positive. Hence for any \(\beta^j \geq 0\), \( \textbf{\emph{H}}=\sum_{j=1}^{N_v}{\beta^j H^j}\) is semi-definite positive \cite{Horn1990}.
\end{IEEEproof}

\section{Proof of Theorem 1}
\begin{IEEEproof}
Since \(\textbf{\emph{K}}\) is a valid kernel and \(\textbf{\emph{H}}\) is semi-definite positive, \(\|f\|_K^2 = \mathbf{f}^T \textbf{\emph{K}} \mathbf{f} = \mathbf{f}^T (\sum_{k=1}^{N_v}{\theta^k K^k})\mathbf{f}=\sum_{k=1}^{N_v}{\theta^k {\|f\|_K^2}_{(k)}}\) and \(\|f\|_I^2 =\mathbf{f}^T \textbf{\emph{H}}\mathbf{f}=\mathbf{f}^T (\sum_{j=1}^{N_v}{\beta^j H^j})\mathbf{f}=\sum_{j=1}^{N_v}{\beta^j {\|f\|_I^2}_{(j)}}\), then  \(\mathcal{G}(\|f\| )=\gamma_A \|f\|_K^2 + \gamma_I \|f\|_I^2\)  is a monotonically increasing real-valued function on \(\|f\|\). According to Lemma 1, the proof of Theorem 1 is complete.
\end{IEEEproof}

\section{Proof of Theorem 2}
\begin{IEEEproof}
Denote the objective function in (\ref{eq4}) as \(F\), the solution at the \(t^{th}\) iteration round as \((f^{(t)},\theta^{(t)},\beta^{(t)})\).

Given fixed \(\theta^{(t-1)}\) and \(\beta^{(t-1)}\) obtained at the \({(t-1)}^{th}\) iteration round, the optimization of (\ref{eq4}) w.r.t. \(f\) at the \(t^{th}\) iteration round degenerates to (\ref{eq1}). This degenerated problem is convex, because the loss function \(\psi\) is convex. Then we have that at the \(t^{th}\) iteration round,
\begin{eqnarray}
F(f^{(t)},\theta^{(t-1)},\beta^{(t-1)}) \leq F(f^{(t-1)},\theta^{(t-1)},\beta^{(t-1)}).\nonumber
\end{eqnarray}

On the other hand, for fixed \(f^{(t)}\) and \(\beta^{(t-1)}\) , the optimization of (\ref{eq4}) w.r.t. \(\theta\) generates to (\ref{eq5}). If \(\textbf{\emph{B}}\) is positive semi-definite, (\ref{eq5}) is a quadratic programming problem. By solving (\ref{eq5}), we have
\begin{eqnarray}
F(f^{(t)},\theta^{(t)},\beta^{(t-1)}) \leq F(f^{(t)},\theta^{(t-1)},\beta^{(t-1)}).\nonumber
\end{eqnarray}

Then for fixed \(f^{(t)}\) and \(\theta^{(t)}\), the minimizing of (\ref{eq4}) w.r.t. \(\beta\) degenerates to (\ref{eq6}) . By solving (\ref{eq6}), we have
\begin{eqnarray}
F(f^{(t)},\theta^{(t)},\beta^{(t)}) \leq F(f^{(t)},\theta^{(t)},\beta^{(t-1)}) \hspace{0.6in} \nonumber\\
\leq F(f^{(t)},\theta^{(t-1)},\beta^{(t-1)}) \leq F(f^{(t-1)},\theta^{(t-1)},\beta^{(t-1)}).\nonumber
\end{eqnarray}
That implies the objective function \(F\) consistently decreases. Therefore, this completes the proof of Theorem 2.

\end{IEEEproof}



\ifCLASSOPTIONcaptionsoff
  \newpage
\fi



\begin{thebibliography}{10}

\bibitem{Bach2004}
Francis R.~Bach, Gert R.~Lanckriet, and Michael I.~Jordan, ``{Multiple kernel learning, conic
  duality, and the SMO algorithm},'' in \emph{Proceedings of the 21st
  International Conference on Machine learning}, Banff, Alberta, Canada, 2004, pp. 41-48.


\bibitem{Belkin2001}
Mikhail~Belkin and Partha~Niyogi, ``{Laplacian eigenmaps and spectral techniques for embedding and
  clustering},'' \emph{Advances in Neural Information Processing Systems},
  Cambridge, MA: MIT Press, 2001, pp. 585-591.

\bibitem{Belkin2006}
Mikhail~Belkin, Partha~Niyogi, and Vikas~Sindhwani, ``{Manifold regularization: A geometric
  framework for learning from labeled and unlabeled examples},'' \emph{Journal
  of Machine Learning Research}, vol.~7, pp. 2399-2434, 2006.


\bibitem{Bezdek2003}
James~C. Bezdek and Richard~J. Hathaway, ``{Convergence of alternating optimization},''
  \emph{Neural, Parallel \& Scientific Computations}, vol.~11, no.~4, pp. 351-368, 2003.

\bibitem{Blum1998}
Avrim~Blum and Tom~Mitchell, ``{Combining labeled and unlabeled data with
  co-training},'' in \emph{Proceedings of the 11th Annual Conference on
  Computational Learning Theory}, Madison, Wisconsin, United States,
  1998, pp. 92-100.



\bibitem{Brefeld2006}
Ulf~Brefeld and Tobias~Scheffer, ``{Semi-supervised learning for structured output
  variables},'' in \emph{Proceedings of the 23rd International Conference on
  Machine Learning}, Pittsburgh, Pennsylvania, 2006, pp. 145-152.

\bibitem{caoliangliang2010a}
Liangliang~Cao, Zicheng~Liu, and Thomas S.~Huang, ``{Cross-dataset action recognition},'' in \emph{Proceedings of the IEEE Conference on Computer Vision and Pattern Recognition}, San Francisco, CA, 2010, pp. 1998-2005.

\bibitem{Caoliangliang2010}
Liangliang~Cao, Jiebo~Luo, Henry Kautz, and Thomas S.~Huang, ``{Image annotation within the context of personal photo collections using hierarchical event and scene models},''  \emph{IEEE Transactons on Multimedia}, vol.~11, no.~2, pp. 208-219, 2009.





\bibitem{Chapelle2005}
Olivier~Chapelle and Alexander~Zien, ``{Semi-supervised classification by low density
  separation},'' in \emph{Proceedings of the International Conference on
Artificial Intelligence and Statistics}, 2005, pp. 57-64.



\bibitem{Collobert2006}
Ronan~Collobert, Fabian~Sinz, Jason~Weston, and L\'{e}on~Bottou, ``{Large scale transductive
  SVMs},'' \emph{Journal of Machine Learning Research}, vol.~7, pp. 1687-1712,
  2006.


\bibitem{Dasgupta2001}
Sanjoy~Dasgupta, Michael L.~Littman, and David~McAllester, ``{PAC generalization bounds for
  co-training},'' \emph{Advances in Neural Information Processing Systems}, Cambridge, MA: MIT Press,
  2001, pp.375-382.

\bibitem{Dempster1977}
A.~P. Dempster, N.~M. Laird, and D.~B. Rubin, ``{Maximum likelihood from
  incomplete data via the EM algorithm},'' \emph{Journal of the Royal
  Statistical Society Series B}, vol.~39, no.~1, pp. 1-38, 1977.

\bibitem{Donoho2003}
David~L. Donoho and Carrie~Grimes, ``{Hessian eigenmaps : new locally linear embedding
  techniques for high-dimensional data},'' in \emph{the National Academy of
  Sciences of the United States of America}, no.~650, pp. 5591-5596, 2003.

\bibitem{Eells1983}
James~Eells and Luc~Lemaire, \emph{{Selected Topics in Harmonic Maps}},\hskip 1em
  plus 0.5em minus 0.4em\relax AMS, Providence, RI, 1983.

\bibitem{Elmoataz2008}
Abderrahim~Elmoataz, Olivier~Lezoray, and S\'{e}bastien~Bougleux, ``{Nonlocal discrete regularization on weighted Graphs: A framework for image and manifold processing},'' \emph{IEEE Transactions on Image Processing},\hskip 1em plus 0.5em minus
  0.4em\relax  vol.~17, no.~7, pp. 1047-1060, 2008.

\bibitem{EveringhamL2007}
Mark~Everingham, Luc~V. Gool, Chris~Williams, John~Winn, and Andrew~Zisserman, ``{the
  PASCAL Visual Object Classes Challenge 2007 (VOC2007)},'' 2007.

\bibitem{Farquhar}
Jason~D.~R. Farquhar, David~R. Hardoon, Hongying~Meng, Sandor~Szedmak, and John~Shawe-taylor,
  ``{Two view learning: SVM-2K, theory and practice},'' \emph{Advances in
  Neural Information Processing Systems},\hskip 1em plus 0.5em minus
  0.4em\relax Cambridge, MA: MIT Press, 2006, pp. 355-362.

\bibitem{Fujino2005}
Akinori~Fujino, Naonori~Ueda, and Kazumi~Saito, ``{A hybrid generative/discriminative
  approach to semi-supervised classifier design},'' in \emph{Proceedings of the
  National Conference on Artifical Intelligence}, 2005, pp. 764-769.



\bibitem{Meh2011}
Mehmet~G\"{o}nen and Ethem~Alpaydm, ``{Multiple kernel learning algorithms},''
  \emph{Journal of Machine Learning Research}, vol.~12, pp. 2211-2268, 2011.

\bibitem{Guillaumin2010}
Matthieu~Guillaumin, Jakob~Verbeek, and Cordelia~Schmid, ``{Multimodal semi-supervised learning for image
  classification},'' in \emph{Proceedings of the IEEE Conference on Computer Vision and Pattern Recognition}, San Francisco, CA, 2010, pp. 902-909.


\bibitem{Horn1990}
 Roger A.~Horn, and  Charles R.~Johnson, \emph{{ Matrix Analysis}}, \hskip 1em plus 0.5em minus
  0.4em\relax Cambridge University Press, 1990.


\bibitem{Jones2005}
Rosie~Jones, ``{Learning to extract entities from labeled and unlabeled text},''
  doctoral disertation, Carnegie Mellon University, 2005.

\bibitem{Kim}
Kwang~In Kim, Florian~Steinke, and Matthias~Hein, ``{Semi-supervised regression using Hessian
  energy with an application to semi-supervised dimensionality reduction},''
\emph{Advances in Neural Information Processing Systems}, Cambridge, MA: MIT Press, 2009, pp.
  979-987.

\bibitem{Lanckriet2004}
Gert R. G.~Lanckriet, Nello~Cristianini, Peter~Bartlett, Laurent El~Ghaoui, and Michael I.~Jordan,
  ``{Learning the kernel matrix with semidefinite programming},'' \emph{Journal
  of Machine Learning Research}, vol.~5, pp. 27-72, 2004.



\bibitem{McFee2011}
Brian~McFee, Carolina~Galleguillos, and Gert~Lanckriet, ``{Contextual object localization with multiple kernel nearest neighbor},'' \emph{IEEE Transactions on Image Processing}, vol.~20, no.~2, pp. 570-585, 2011.

\bibitem{Miller1996}
David J.~Miller and Hasan S.~Uyar, ``{A mixture of experts classifier with learning
  based on both labelled and unlabelled Data},'' \emph{Advances in Neural
  Information Processing Systems}, Cambridge, MA: MIT Press, 1996, pp. 571-577.

\bibitem{Nesterov2005a}
Yu~Nesterov, ``{Smooth minimization of non-smooth functions},''
  \emph{Mathematical Programming}, vol.~103, no.~1, pp. 127-152, 2005.

\bibitem{Nigam2000}
Kamal~Nigam and Rayid~Ghani, ``{Analyzing the effectiveness and applicability of
  co-training},'' in \emph{Proceedings of the 9th International Conference on
  Information and Knowledge Management}, 2000, pp. 86-93.

\bibitem{KamalNi}
Kamal~Nigam, Andrew~kachites Maccallum, Sebastian~Thrun, and Tom~Mitchell, ``{Text
  classification from labeled and unlabeled documents using EM},''
  \emph{Machine Learning}, vol.~39, no.~2, pp. 103-134, 2000.

\bibitem{Nilufar}
 Sharmin~Nilufar, Nilanjan~Ray, and Hong~Zhang, ``{Object detection with DoG scale-space: A multiple kernel learning approach}'', \emph{IEEE Transactions on Image Processing}, 03 April 2012.






\bibitem{Sonnenburg2006}
S\"{o}ren~Sonnenburg, Gunnar~R\"{a}tsch, Christin~Sch\"{a}fer, and Bernhard~Sch\"{o}lkopf, ``{Large
  scale multiple kernel learning},'' \emph{Journal of Machine Learning
  Research}, vol.~7, pp. 1531-1565, 2006.

\bibitem{Steinke2008}
Florian~Steinke and Matthias~Hein, ``{Non-parametric regression between manifolds},''
\emph{Advances in Neural Information Processing Systems}, Cambridge, MA: MIT Press 2008, pp.
  1561-1568.

\bibitem{Tian2008}
Xinmei~Tian, Linjun~Yang, Jingdong~Wang, Xiuqing Wu, and Xian-sheng Hua, ``{Bayesian visual reranking},''  \emph{IEEE Transactions on Multimedia}, vol.~13, no.~4, pp. 639-652, 2011.

\bibitem{Vapnic1998}
Vladimir N.~Vapnic, \emph{{Statistical Learning Theory}},\hskip 1em plus 0.5em minus
  0.4em\relax Wiley, 1998.

\bibitem{Wangmeng2007}
Meng~Wang, Xian-Sheng~Hua, Richang~Hong, Jinhui~Tang, Guo-Jun~Qi, and Yan~Song, ``{Unified video annotation via multigraph learning},'' \emph{IEEE Transactions on Circuits and systems for video technology}, vol.~19, no.~5, pp. 733-346, 2009.

\bibitem{Wangmeng2009}
Meng~Wang, Kuiyuan~Yang, Xian-Sheng~Hua, and Hong-Jiang Zhang, ``{Towards relevant and diverse search of social images},'' \emph{IEEE Transactions on Multimedia}, vol.~12, no.~8 pp. 829-842, 2010.


\bibitem{Xia2010a}
Tian~Xia, Dacheng~Tao, Tao~Mei, and Yongdong~Zhang, ``{Multiview Spectral Embedding},''
  \emph{IEEE Transactions
  on Systems, Man, and Cybernetics, Part B: Cybernetics}, vol.~40, no.~6, pp. 1438-1446, 2010.

\bibitem{Xie2011}
Bo~Xie, Yang~Mu, Dacheng~Tao, and Kaiqi~Huang, ``{m-SNE: multiview stochastic neighbor
  embedding},'' \emph{IEEE
  Transactions on Systems, Man, and Cybernetics, Part B: Cybernetics}, vol.~41, no.~4, pp. 1088-1096, 2011.



\bibitem{Zhu2007}
Xiaojin~Zhu, ``{Semi-Supervised Learning Literature Survey Contents},'' Technical Report 1530, Computer Sciences, University of Wisconsin-Madison, 2005.


\end{thebibliography}
\end{document}